\documentclass[11pt,a4paper]{article}
\usepackage[hyperref]{eacl2021}
\usepackage{url}
\usepackage{times}
\usepackage{latexsym}
\usepackage{enumitem}
\usepackage{array,multirow}
\usepackage{tabularx}
\newcolumntype{L}{>{\centering\arraybackslash}m{13cm}}
\usepackage{microtype}
\usepackage{booktabs}
\newcommand{\cc}[1]{\multicolumn{1}{c}{#1}}
\defcitealias{Davidson2017}{Davidson}
\defcitealias{Mandl2019}{Mandl}
\makeatletter
\def\blfootnote{\xdef\@thefnmark{}\@footnotetext}
\makeatother
\usepackage{graphicx}

\aclfinalcopy
\def\aclpaperid{9}
\def\confidential{\textcolor{black}{WASSA 2021 Submission~\aclpaperid.  Confidential Review Copy.  DO NOT DISTRIBUTE.}}

\title{Hate Towards the Political Opponent:\\
  A Twitter Corpus Study of the 2020 US Elections on the\\ Basis of
  Offensive Speech and Stance Detection}

\author{Lara Grimminger \and Roman Klinger \\
  Institut f\"ur Maschinelle Sprachverarbeitung, University of Stuttgart \\
  Pfaffenwaldring 5b, 70569 Stuttgart, Germany\\
  \{lara.grimminger,roman.klinger\}@ims.uni-stuttgart.de \\
}

\date{}

\begin{document}
\maketitle
\begin{abstract}
  The 2020 US Elections have been, more than ever before,
  characterized by social media campaigns and mutual accusations. We
  investigate in this paper if this manifests also in online
  communication of the supporters of the candidates Biden and Trump,
  by uttering hateful and offensive communication.  We formulate an
  annotation task, in which we join the tasks of hateful/offensive
  speech detection and stance detection, and annotate 3000 Tweets from
  the campaign period, if they express a particular stance towards a
  candidate. Next to the established classes of \emph{favorable} and
  \emph{against}, we add \emph{mixed} and \emph{neutral} stances and
  also annotate if a candidate is mentioned without an opinion
  expression. Further, we annotate if the tweet is written in an
  offensive style. This enables us to analyze if supporters of Joe
  Biden and the Democratic Party communicate differently than
  supporters of Donald Trump and the Republican Party. A BERT baseline
  classifier shows that the detection if somebody is a supporter of a
  candidate can be performed with high quality (.89 F$_1$ for Trump
  and .91 F$_1$ for Biden), while the detection that somebody
  expresses to be against a candidate is more challenging (.79 F$_1$
  and .64 F$_1$, respectively). The automatic detection of
  hate/offensive speech remains challenging (with .53 F$_1$). Our
  corpus is publicly available and constitutes a novel resource for
  computational modelling of offensive language under consideration of
  stances.
\end{abstract}

\blfootnote{\hspace{-0.65cm} This paper contains offensive language.}

\section{Introduction}
Social media are indispensable to political campaigns ever since
Barack Obama used them so successfully in 2008
\citep{Tumasjan2010}. Twitter in particular is a
much-frequented form of communication with monthly 330 million active
users \citep{Clement2019}. The microblogging platform was credited to
have played a key role in Donald Trump's rise to power
\citep{Stolee2018}. As Twitter enables users to express their
opinions about topics and targets, the insights gained from detecting
stance in political tweets can help monitor the voting base.

In addition to the heated election of Trump in 2016, the world has
also seen an increase of hate speech \citep{GaoHuang2017}. Defined as
``any communication that disparages a target group of people based on
some characteristic such as race, colour, ethnicity, gender, sexual
orientation, nationality, religion, or other characteristic''
\citep{Nockelby2000}, hate speech is considered ``a particular form of
offensive language'' \citep{Warner2012}. However, some authors also
conflate hateful and offensive speech and define hate speech as
explicitly or implicitly degrading a person or group
\citep{GaoHuang2017}. Over the years, the use of hate speech in social
media has increased \citep{DeGibert2018}. Consequently, there is a
growing need for approaches that detect hate speech automatically
\citep{GaoHuang2017}.

From the perspective of natural language processing (NLP), the
combination of political stance and hate speech detection provides
promising classification tasks, namely determining the attitude a text
displays towards a pre-determined target and the presence of hateful
and offensive speech. In contrast to prior work on stance detection
\cite[i.a.]{Somasundaran2010,Mohammad2016}, we not only annotate if a
text is \emph{favorable}, \emph{against} or does not mention the
target at all (\emph{neither}), but include whether the text of the
tweet displays a \emph{mixed} (both favorable and against) or
\emph{neutral} stance towards the targets. With this formulation we
are also able to mark tweets that mention a target without taking a
clear stance. To annotate hateful and offensive tweets, we follow the
definition of \newcite{GaoHuang2017} and adapt our guidelines to
political discourse.

Our contributions are the following:
\begin{itemize}
\item We publish a Twitter-corpus that is annotated both for stance
  and hate speech detection. We make this corpus of 3000 Tweets
  publicly available at
  \url{https://www.ims.uni-stuttgart.de/data/stance_hof_us2020}.
\item Based on a manual analysis of these annotations, our results
  suggest that Tweets that express a stance against Biden contain more
  hate speech than those against Trump.
\item Our baseline classification experiments show that the detection
  of the stance that somebody is in-favor of a candidate performs
  better than that somebody is against a candidate. Further, the
  detection of hate/offensive speech on this corpus remains
  challenging.
\end{itemize}

\section{Related Work}
\subsection{Hate Speech and Offensive Language}

In early work on hate speech detection, \citet{Spertus1997} described
various approaches to detect abusive and hostile messages occurring
during online communication. More recent work also considered
cyberbullying \citep{Dinakar2012} and focused on the use of
stereotypes in harmful messages \citep{Warner2012}. Most of the
existing hate speech detection models are supervised learning
approaches. \citet{Davidson2017} created a data set by collecting
tweets that contained hate speech keywords from a crowd-sourced hate
speech lexicon. They then categorized these tweets into hate speech,
offensive language, and neither. \citet{Mandl2019} sampled their data
from Twitter and partially from Facebook and experimented with binary
as well as more fine-grained multi-class classifications. Their results
suggest that systems based on deep neural networks performed best.

\newcite{Waseem2016} used a feature-based approach to explore several
feature types. \citet{Burnap2014} collected hateful tweets related to
the murder of Drummer Lee Rigby in 2013. The authors examined
different classification methods with various features including
n-grams, restricted n-grams, typed dependencies, and hateful
terms. \citet{Schmidt2017} outlined that the lack of a benchmark data
set based on a commonly accepted definition of hate speech is
challenging. \citet{Ross2016} found that there is low agreement among
users when identifying hateful messages.

For the SemEval 2019 Task 5, \citet{Basile2019} proposed two hate
speech detection tasks on Spanish and English tweets which contained
hateful messages against women and immigrants. Next to a binary
classification, participating systems had to extract further features
in harmful messages such as target identification. None of the
submissions for the more fine-grained classification task in English
could outperform the baseline of the task organizers. In case of Spanish,
the best results were achieved by a linear-kernel SVM. The authors
found that it was harder to detect further features than the presence
of hate speech. The recent shared task on offensive language
identification organized by \citet{Zampieri2020} was featured in five
languages. For a more detailed overview, we refer to the
surveys by \newcite{Mladenovic2021,Fortuna2018,Schmidt2017}.

In contrast to this previous work, we provide data for a specific
recent use case, and predefine two targets of interest to be analyzed.

\subsection{Stance Detection}
Related work on stance detection includes stance detection on
congressional debates \citep{Thomas2006}, online forums
\citep{Somasundaran2010}, Twitter \citep{Mohammad2016, Mohammad2017,
  Aker2017, Can2018, Lozknikov2020} and comments on news
\citep{Lozknikov2020}. \citet{Thomas2006} used a corpus of speeches
from the US Congress and modeled their support/oppose towards a
proposed legislation task. \citet{Somasundaran2010} conducted
experiments with sentiment and arguing expressions and used features
based on modal verbs and sentiments for stance classification. For the
SemEval 2016 Task 6 organized by \citet{Mohammad2016}, stance was
detected from tweets. The task contained two stance detection subtasks
for supervised and weakly supervised settings. In both classification
tasks, tweet-target pairs needed to be classified as either Favor,
Against or Neither. The baseline of the task organizers outperformed
all systems’ results that were submitted by task participants.

In hope that sentiment features would have the same effect on stance
detection as they have on sentiment prediction, \citet{Mohammad2017}
concurrently annotated a set of tweets for both stance and
sentiment. Although sentiment labels proved to be beneficial for
stance detection, they were not sufficient. Instead of a
target-specific stance classification, \citet{Aker2017} described an
open stance classification approach to identify rumors on Twitter.
The authors experimented with different classifiers and task-specific
features which measured the level of confidence in a tweet. With the
additional features, their approach outperformed state-of-the-art
results on two benchmark sets.

In addition to this previous work, we opted for a more fine-grained
stance detection and not only annotated \textit{favor},
\textit{against} and \textit{neither} towards a target but also
whether the stance of the text was \textit{mixed} or
\textit{neutral}. Further, we combine stance detection with
hate/offensive speech detection.

\section{Corpus}
\subsection{Data Collection}

Our goal is on the one side to create a new Twitter data set that
combines stance and hate/offensive speech detection in the political
domain. On the other side, we create this corpus to investigate the
question how hate/offensive speech is distributed among different
stances.

We used the Twitter API v 1.1.\ to fetch tweets for 6 weeks leading to
the presidential election, on the election day and for 1 week after
the election. As search terms, we use the mention of the presidential
and vice presidential candidates and the outsider West; the mention of
hashtags that show a voter's alignment such as the campaign slogans of
the candidate websites, and further nicknames of the candidates. The
list of search terms is: \#Trump2020, \#TrumpPence2020, \#Biden2020,
\#BidenHarris2020, \#Kanye2020, \#MAGA2020,
\#BattleForTheSoulOfTheNation, \#2020Vision, \#VoteRed2020,
\#VoteBlue2020, Trump, Pence, Biden, Harris, Kanye, President, Sleepy
Joe, Slow Joe, Phony Kamala, Monster Kamala.

After removing duplicate tweets, the final corpus consists of 382.210
tweets. From these, there are 220.941 that contain Trump
related hashtags and mentions, 230.629 tweets that carry hashtags and
mentions associated with Biden and 1.412 tweets with hashtags and
mentions related to Kanye West.

\subsection{Annotation}
\subsubsection{Annotation Task}
From the 382.210 tweets, we sampled 3000 tweets for annotation. Given
the text of a tweet, we rated the stance towards the
targets Trump, Biden, and West in the text. The detected stance can be
from one of the following labels:
\begin{itemize}
\item \emph{Favor}: Text argues in favor of the target
\item \emph{Against}: Text argues against the target
\item \emph{Neither}: Target is not mentioned; neither implicitly nor explicitly
\item \emph{Mixed}: Text mentions positive as well as negative aspects about the target
\item \emph{Neutral}: Text states facts or recites quotes; unclear, whether text holds any position towards the target.
\end{itemize}
The default value shown in the annotation environment is \emph{Neither}.

The text was further annotated as being hateful and
non-hateful. We did not separate if a group or a single person was
targeted by hateful language. Further, we adapted the guidelines on
hate speech annotation to be able to react to name-calling and down
talking of the political opponent. Thus, we rated expressions such as
``Dementia Joe'' and ``DonTheCon'' as hateful/offensive (HOF).

\begin{table}[t]
  \centering
  \begin{tabular}{l rrrr}
    \toprule
    & \multicolumn{3}{c}{Stance} & \\
    \cmidrule(lr){2-4}
    Iteration &Trump&Biden&West&HOF \\
    \cmidrule(r){1-1}\cmidrule(lr){2-2}\cmidrule(lr){3-3}\cmidrule(lr){4-4}\cmidrule(l){5-5}
    1 A1+A2& 0.83 & 0.81  & 0.00 & $-$0.02 \\
    2 A1+A2& 0.78 & 0.78 & 0.75 & 0.42 \\
    3 A1+A2& 0.81  & 0.88 & 0.00 & 0.73 \\
    4 A2+A3& 0.61 & 0.76  & 0.75 & 0.62 \\
    \bottomrule
  \end{tabular}
  \caption{Cohen’s $\kappa$ for stance and hate/offensive speech (HOF).}
  \label{iaa}
\end{table}

\subsubsection{Annotation Procedure}
To evaluate the annotation guidelines (which we make available
together with the data) we perform multiple annotation iterations with
three annotators. Annotator 1 is a 22 year old male undergraduate student
of computational linguistics who speaks German, English, Catalan, and
Spanish. Annotator~2 is a 26 year old female undergraduate student of
computational linguistics who speaks German and English. Annotator 3
is a 29 year old female graduate student of computational
linguistics who speaks German and English. Annotator 1 and 2 annotated
300 tweets in three iterations with 100 tweets per iteration. After
each iteration, the annotators discussed the tweets they rated
differently and complemented the existing guidelines. Finally,
Annotator 2 and 3 annotated 100 tweets with the improved guidelines to
check whether the rules are clear and understandable, especially if
read for the first time.

\begin{table}
  \centering
  \setlength{\tabcolsep}{7pt}
  \begin{tabular}{lrrrr}
    \toprule
    Class &  \cc{HOF} & \cc{$\neg$HOF} &  \cc{$\sum$} & \cc{\%HOF} \\
    \cmidrule(r){1-1}\cmidrule(rl){2-3}\cmidrule(rl){4-4}\cmidrule(rl){5-5}
    Favor   &      101 &          679 &         780 & 12.9 \\
    Against &      156 &          686 &         842 & 18.5\\
    Neither &       76 &          941 &        1017 &  7.5 \\
    Mixed   &        1 &           19 &          20 &  5.0 \\
    Neutral &       18 &          323 &         341 &  5.3 \\
    \cmidrule(r){1-1}\cmidrule(rl){2-3}\cmidrule(rl){4-4}\cmidrule(rl){5-5}
    Total      &      352 &         2648 &        3000 & 11.7 \\
    \bottomrule
  \end{tabular}
  \caption{Distribution of tweets about target Trump}
  \label{trump_hate}
\end{table}

Table~\ref{iaa} shows the result of Cohen’s $\kappa$ of each
iteration. In the first iteration, the agreement for HOF is purely
random ($-0.02\kappa$), the stance annotations show acceptable
agreement (.83, .81$\kappa$, respectively for Trump and Biden). West
has not been mentioned in any of the 100 tweets. In a group discussion
to identify the reasons for the substantial lack of agreement for HOF,
we developed guidelines which described hateful and offensive speech
in more detail and added further examples to our guidelines. We
particularly stressed to annotate name-calling as hateful and
offensive. This showed success in a second iteration with .42$\kappa$
for HOF agreement. The scores for Trump and Biden decreased slightly
but still represented substantial agreement. We carried out another
group discussion to discuss tweets where Annotator 1 and 2 chose
different classes. We particularly refined the guidelines for class
\emph{Neutral mentions} and included offensive and hateful
abbreviations such as ``POS'' (``piece of shit'') and ``BS''
(``bullshit'') which have been missed before. This led to a HOF
agreement of .73$\kappa$, while the stance agreement remained on
similar levels (.81, .88).

As a concluding step, Annotator 2 and 3 rated 100 tweets. The
annotators were provided with the guidelines established during the
iterations between Annotator 1 and 2. Table~\ref{iaa} shows that the
inter-annotator agreement for HOF is 0.62, for target Trump 0.61, for
target Biden 0.76 and for target West 0.75. These scores indicate
substantial agreement between Annotator 2 and 3 based on comprehensive
guidelines. The final annotation of the overall data set has been
performed by Annotator~2.

\begin{table}
  \centering
  \setlength{\tabcolsep}{7pt}
  \begin{tabular}{lrrrr}
    \toprule
    Class &  \cc{HOF} & \cc{$\neg$HOF} &  \cc{$\sum$} & \cc{\%HOF} \\
    \cmidrule(r){1-1}\cmidrule(rl){2-3}\cmidrule(rl){4-4}\cmidrule(rl){5-5}
    Favor   &      141 &         1095 &        1236 & 11.4 \\
    Against &      108 &          296 &         404 & 26.7 \\
    Neither &       87 &          900 &         987 & 8.8 \\
    Mixed   &        6 &           41 &          47 & 12.8 \\
    Neutral &       10 &          316 &         326 &  3.1  \\
    \cmidrule(r){1-1}\cmidrule(rl){2-3}\cmidrule(rl){4-4}\cmidrule(rl){5-5}
    Total   &      352 &         2648 &        3000 & 11.7 \\
    \bottomrule
  \end{tabular}
  \caption{Distribution of tweets about target Biden}
  \label{biden_hate}
\end{table}
 
\section{Results}
\subsection{Corpus Statistics}
We now analyze the corpus for the targets Trump and Biden to answer
the question if supporters of Trump (and Pence) use more hateful and
offensive speech than supporters of Biden (and
Harris). Tables~\ref{trump_hate} and~\ref{biden_hate} show the
distribution of the classes \emph{Favor, Against, Neither, Mixed,
  Neutral mentions} and how often each class was labeled as HOF or
Non-HOF ($\neg$HOF).

The data set is unbalanced: only 11.7\% of the tweets are
hateful/offensive. Furthermore, there are more tweets labeled as
\emph{Favor, Against}, and \emph{Neither} than \emph{Mixed}, or
\emph{Neutral mentions} for target Trump. In case of target Biden,
more tweets are labeled as \emph{Favor} and as \emph{Neither} than as
\emph{Against, Mixed}, or \emph{Neutral mentions}. In total, there
were only 9 tweets about Kanye West in the annotated data set, which
is why we do not present statistics about him.

\textbf{Did Trump supporters use more hateful and offensive speech
  than supporters of Biden?} A comparison of Tables~\ref{trump_hate}
and~\ref{biden_hate} suggests that supporters of team Trump use
slightly more often harmful and offensive speech with 12.9\% than
supporters of team Biden, with 11.4\%. This indicates that Trump
supporters use more hateful speech than supporters of Biden, yet, this
difference is only minor.  This is arguable a result of the aspect
that HOF is also often expressed without naming the target
explicitly. Furthermore, given the fact that we added offensive
nicknames such as ``Sleepy Joe'' to our search terms, this result is
biased.

By means of pointwise mutual information we identified the top
10 words that are unlikely to occur in a tweet labeled as
Non-Hateful. As Table~\ref{pmi:hof} shows, these words are offensive
and promote hate. This list also mirrors the limitations of our search
terms as the adjective "sleepy" is part of the top 10.

Likewise, we identified the top 10 words that are unlikely to appear
in a tweet labeled as \emph{Favor} towards Trump and thus, argue
against him. Next to hashtags that express a political preference for
Biden, the top 10 list contains words that refer to Trump's taxes and
a demand to vote. Similarly, the top 10 words that are unlikely to
occur in a tweet labeled as \emph{Favor}ing Biden and therefore
express the stance \emph{Against} him, consist of adjectives Trump
mocked him with (creepy, sleepy) as well as a reference to his son
Hunter.

\begin{table}[t]
  \centering
   \setlength{\tabcolsep}{7pt}
  \begin{tabular}{rrr}
    \toprule
    \cc{HOF} & \cc{Against Trump} & \cc{Against Biden} \\
     \cmidrule(r){1-1}\cmidrule(rl){2-2}\cmidrule(r){3-3}
    fuck & \#biden2020 & \#trump2020 \\ disgusting & \#bidenharris2020 & @seanhannity \\ idiots & \#trumptaxreturns & hunter \\ bitch & taxes & creepy \\ fat& \#vote & top \\ bullshit & very &sleepy \\ vs & gop & radical \\idiot & pence & evidence\\ \#covidiot & vote. & leading \\sleepy & thinks & across\\
    \bottomrule
  \end{tabular}
  \caption{Results of the pointwise mutual information calculation}
  \label{pmi:hof}
\end{table}

\textbf{Who is more targeted by hateful and offensive speech, Biden
  and the Democratic party or Trump and the Republican Party?} We note
that 26.7\% of the tweets against target Biden contain
hateful/offensive language, whereas only 18.5\% of the tweets against
target Trump are hateful/offensive. Thus, our results suggest that
Biden and the Democratic Party are more often targets of hateful and
offensive tweets than Trump and the Republican Party.

However, from this analysis we cannot draw that the offensive stems
from supporters of the other party. Due to the limitations in our
search terms we also note that there might be an unknown correlation
of the search terms to HOF which we cannot entirely avoid. Further,
these results should be interpreted with a grain of salt, given that
the sampling procedure of the Twitter API is not entirely transparent.

\begin{table}
  \centering
  \setlength{\tabcolsep}{7pt}
  \begin{tabular}{l ccc ccc}
    \toprule
   & \multicolumn{3}{c}{Target Trump} &  \multicolumn{3}{c}{Target Biden} \\
    \cmidrule(r){2-4}\cmidrule(l){5-7}
    Class &P&R&F$_1$&P&R&F$_1$\\
    \cmidrule(lr){1-1}\cmidrule(lr){2-4}\cmidrule(lr){5-7}
    Against & .77 & .81 & .79 & .67 & .62 & .64\\
    Favor & .88 & .90 & .89 & .90 & .93 & .91\\
    Mixed &  .00 & .00 & .00 & .00 & .00 & .00\\
    Neither & .95 & .95 & .95 & .93 & .99 & .96\\
    Neutral & .58 & .49 & .53 & .59 & .58 & .59 \\
    \bottomrule
  \end{tabular}
  \caption{Precision, Recall, and F$_1$ of stance detection baseline for targets Trump and Biden}
  \label{stance_detection}
\end{table}

\begin{table*}
  \centering
  \setlength{\tabcolsep}{8pt}
  \renewcommand\arraystretch{1.20}
  \begin{tabular}{c c c ccc ccc ccc}
    \toprule
    &&&\multicolumn{9}{c}{Test data}\\
    \cmidrule{4-6}\cmidrule{7-9}\cmidrule{10-12}
    &&& \multicolumn{3}{c}{\citetalias{Davidson2017}} & \multicolumn{3}{c}{\citetalias{Mandl2019}} &  \multicolumn{3}{c}{Ours}\\
    \cmidrule(lr){4-6}\cmidrule(lr){7-9}\cmidrule(lr){10-12}
    & &Class &P&R&F$_1$ &P&R&F$_1$ &P&R&F$_1$\\
    \cmidrule(r){3-3}\cmidrule(lr){4-6}\cmidrule(lr){7-9}\cmidrule(lr){10-12}
    \multirow{6}{*}{\rotatebox[origin=c]{90}{\centering Train data}}
    & \multirow{2}{*}{\citetalias{Davidson2017}}
    & HOF & .98 & .98 & .98 & .34 & .55 & .42 & .08 & .75 & .14 \\
    && Non-HOF & .89 & .88 & .88 & .48 & .28 & .36 & .71 & .07 & .12\\[1mm]
    \cmidrule(r){3-3}\cmidrule(lr){4-6}\cmidrule(lr){7-9}\cmidrule(lr){10-12}
    & \multirow{2}{*}{\citetalias{Mandl2019}}
      & HOF & .61 & .24 & .35 & .66 & .48 & .56 & .20 & .53 & .29 \\
    && Non-HOF & .06 & .22 & .09 & .70 & .83 & .76 & .94 & .77 & .85\\[1mm]
    \cmidrule(r){3-3}\cmidrule(lr){4-6}\cmidrule(lr){7-9}\cmidrule(lr){10-12}
    & \multirow{2}{*}{\rotatebox[origin=c]{0}{\parbox[c]{1cm}{\centering Ours}}} 
      & HOF & .52 & .16 & .25 & .52 & .60 & .56 & .52 & .54 & .53 \\
    && Non-HOF & .06 & .25 & .09 & .70 & .62 & .66 & .95 & .94 & .95\\
    \bottomrule
  \end{tabular}
  \caption{F1 scores of the hate speech detection baseline model trained and tested on different corpora}
  \label{hate_detection}
\end{table*}

\subsection{Stance Classification Experiments}
\label{stanceresults}

Next to the goal to better understand the distribution of
hate/offensive speech during the election in 2020, the data set
constitutes an interesting resource valuable for the development of
automatic detection systems. To support such development, we provide
results of a baseline classifier. We used the pretrained BERT base
model\footnote{\url{https://huggingface.co/bert-base-uncased}}
\cite{Devlin2019} and its TensorFlow implementation provided by
HuggingFace\footnote{\url{https://github.com/huggingface/transformers}}
\cite{Wolf2020}. Our data set was divided into 80\% for training and
20\% for testing.

Each model was trained with a batch size of 16, a learning rate (Adam)
of $5\cdot10^{-5}$, a decay of 0.01, a maximal sentence length of 100
and a validation split of 0.2. Further, we set the number of epochs to
10 and saved the best model on the validation set for testing.

Table~\ref{stance_detection} shows the results for stance detection
prediction. We observe that not all classes can be predicted equally
well. The two best predicted classes for Trump are \emph{Neither} and
\emph{Favor} with a F$_1$ score of 0.95 and 0.89, respectively. These
scores are followed by class \emph{Against} with a F$_1$ score of
0.79. However, our model had difficulties to correctly predict the
class \emph{Neutral}, with a more limited precision and recall (.58
and .49). The class \emph{Mixed} could not be predicted at all.

These results only partially resemble for the target Biden: The
classes \emph{Neither} and \emph{Favor} have the highest F$_1$ score
with 0.96 and 0.91, respectively. In contrast to target Trump, the
performance of our model to predict the class \emph{Against} is much
lower (.64 F$_1$). The F$_1$ score of class \emph{Neutral} is low
again with .59 and class \emph{Mixed} could not be predicted. We
conclude that stance can be detected from tweets. Yet, our results
suggest that it is more challenging to predict fine-grained stance
classes such as \emph{Mixed} and \emph{Neutral mentions} than the
classes \emph{Favor, Against} and \emph{Neither}.
This result is, at least partially, a consequence of the
data distribution. The \emph{Mixed} label has very few instances
(20+47); the \emph{Neutral} label is the second most seldomly
annotated class, though it is substantially more frequent (341+326).

\begin{table*}[t]
  \centering
  \renewcommand\arraystretch{1.05}
  \setlength{\tabcolsep}{5pt}
  \begin{tabularx}{\linewidth}{lccX}
    \toprule
    Target &Pred&Gold&Text \\
    \cmidrule(lr){1-1}\cmidrule(lr){2-3}\cmidrule(lr){4-4}
    Trump & A & A & TWO HUNDRED THOUSAND PEOPLE HAVE DIED OF \#COVID19 UNDER Trump's WATCH. IT DID NOT HAVE TO BE LIKE THIS. \#BidenHarris2020 will take steps to make us safe. Trump is happy to let us burn, and so he is a \#weakloser \#VoteBidenHarrisToSaveAmerica \#RepublicansForBiden \\
    Trump & F & F & Trump is making all types of economic/peace deals. While Democrats are creating mobs tearing down historical statutes and destroying WHOLE cities. It is a NO brainer on who to vote for in 2020. Trump builds! Democrats DESTROY! \#Trump2020. \#Trump2020LandslideVictory\\
    Trump & A & F & President Trump please don't let this son of a bitch crazy creepy pedophile motherfucker of Joe Biden and his brown paper bag bitch of Kamala Harris win this election do not let them win. \\
    Trump & F & A & \#Trump2020? Wish full recovery to see him ask forgiveness to the country for his incompetence and lack of respect for the American people during Covid-19 crisis.\#VictoryRoad to \#Biden/Harris \\
    Biden & A & A & Joe Biden is a weak weak man in many ways. Jimmy Carter by half. 1/2 of America literally can't stand Kamala today. Women will hate on her viciously.  Enjoy the shit sandwich.\\
    Biden & F & F & Kamala was my first choice, but I agree at this moment in time. Joe Biden is the right choice. We are lucky that he is willing to continue to serve our country. When I voted for Biden/Harris I felt good, I felt hopeful, I know this is the right team to recover our country/democracy \\
    Biden & A & F & @KamalaHarris @JoeBiden Dominate and Annihilate trump, Joe aka 46 \#BidenHarris2020 \#WinningTeam \#PresidentialDebate \#TrumpTaxReturns \#TrumpHatesOurMilitary \#TrumpKnew \#TrumpLiedPeopleDied GO JOE\\
    Biden & F & A & While we are here, this type of BS is what Kamala Harris and Joe Biden call "science". \\
    \bottomrule
  \end{tabularx}
  \caption{Examples of correct and incorrect predictions of favor (F) and against (A) stance in the tweets.}
  \label{analysis_stance}
\end{table*}

\subsection{Cross-Corpus Hate Speech Detection Experiments}
Similar to the stance detection baseline results, we now report
results of a classifier (configured the same as the one in
Section~\ref{stanceresults}). To obtain an understanding how challenging
the prediction is on our corpus, and how different the concept of
hate/offensive speech is from existing resources, we perform this
analysis across a set of corpora, as well as inside of each corpus.

To that end, we chose the following hate speech/offensive speech corpora:

\begin{enumerate}
\item Data Set 1 by \newcite{Davidson2017}.\\ This corpus contains
  24.783 tweets, categorized into into hateful, offensive, and
  neither. In our study, we we only use two classes, hateful/offensive
  and non-hateful. Therefore, we conflate the two classes, hateful and
  offensive, into one. We randomly split their data, available at
  \url{https://github.com/t-davidson/hate-speech-and-offensive-language},
  into 80\% for training and 20\% for testing.
\item Data Set 2 by \newcite{Mandl2019}.\\ In this study, the authors
  conducted three classification experiments including a binary one,
  where 5.852 posts from Twitter and Facebook were classified into
  hate speech and non-offensive (Sub-task A). From their multi-lingual
  resource, we only need the English subset. We use the training data
  available at
  \url{https://hasocfire.github.io/hasoc/2019/dataset.html} and
  perform a 80/20\% train/test split.
\end{enumerate}

Table~\ref{hate_detection} shows the results for all combinations of
training on the data by \newcite{Davidson2017}, \newcite{Mandl2019},
and ours (presented in this paper). When we only look at the results
of the model when trained and tested on subcorpora from the same
original source, we observe that there are some noteworthy
differences. The recognition of HOF on the data by
\newcite{Davidson2017} shows a high .98 F$_1$ (note that this result
cannot be compared to their original results, because we conflate two
classes). Training and testing our baseline model on the data by
\newcite{Mandl2019} shows .56 F$_1$ but with a particularly limited
recall. On our corpus, given that it is the smallest one, the model
performs still comparably well with .53 F$_1$. Precision and recall
values are more balanced for the other corpora than for
\newcite{Mandl2019}. Note that these results are comparably low in
comparison to other previously published classification
approaches. However, they allow for a comparison of the performances
between the different corpora. We particularly observe that the data
set size seems to have an impact on the predictive performance.

When we move to a comparison of models trained on one corpus and
tested on another, we see that the subcorpora created for a binary
classification experiment yield better results. The imbalance of
labels caused by the conflation of two classes on the data by
\newcite{Davidson2017} led to weak predictions on the other
subcorpora.

Therefore, we conclude that the concept of hate/offensive speech
between these different resources is not fully comparable, be it due
to different instances, settings or annotators. The development of
models that generalize across domains, corpora, and annotation
guidelines is challenging.

\begin{table*}
  \centering
  \setlength{\tabcolsep}{5pt}
  \begin{tabularx}{\linewidth}{llX}
    \toprule
     Pred&Gold&Text \\
    \cmidrule(r){1-1}\cmidrule(rl){2-2}\cmidrule(lr){3-3}
    HOF & HOF & Two Kamala Harris staffers have covid-19. Let's hope at least one of them has been recently sniffed by Creepy Joe.\\
    HOF & $\neg$HOF & He's a badass!!! \#Trump2020 \#Suckit \#Winning\\
    $\neg$HOF & HOF & The democrats are literally the nazis. If they pack the courts and pass the 25th amendment Joe Biden and Kamala Harris will be in the exact same place that hindenburg and hitler were in. The 25th amendment is almost the same exact law hitler got passed in order to take power.
 \\
    \bottomrule
  \end{tabularx}
  \caption{Examples of correct and incorrect predictions for hateful and offensive speech in the tweets.}
  \label{analysis_hate}
\end{table*}

\subsection{Analysis}
We now take a closer look at the tweets and their predicted classes to
explore why tweets have been misclassified. We show examples in
Table~\ref{analysis_stance} for stance classification.

Our model performed well when predicting the class \emph{Favor} for
both targets. The examples in Table~\ref{analysis_stance} show a
common pattern, namely that tweets being in favor of the respective
target praise target's achievements and contain words of support such
as \emph{builds, vote} and \emph{right choice}. Additionally, users
often complement their tweets with target-related hashtags, including
\emph{Trump2020, Trump2020LandslideVictory} and \emph{Biden/Harris} to
stress their political preference. However, these hashtags can be
misleading as they not always express support of the candidate. The
4th example contains the hashtag \emph{\#Trump2020} and was therefore
predicted to be in favor of Trump, while it actually argues against
him. In the 5th example, the irony expressed by the quotation marks
placed around the word \emph{science} and the offensive expression
\emph{BS} for ``bullshit'' were not detected.

Supporters of both candidates verbally attack each other over who to
vote for and use hashtags and expressions to make the opposite side
look poorly. Looking at tweets incorrectly labeled as \emph{Against},
we see that in case of target Trump the string of insults addressing
Biden and Harris possibly confused our baseline and led to a
misclassification of the tweet. Turning to Biden, the sentence
\emph{Joe aka 46} was not detected to be positive and supportive.

We also show a set of examples for hate/offensive speech detection in
Table~\ref{analysis_hate}. As the first tweet exemplifies, tweets
correctly predicted as HOF often contain one or more hate and
offensive key words, e.g. \emph{Creepy Joe}. The first example also
wishes Joe Biden to fall ill with Covid-19.

However, although the 2nd example seems to contain offensive words
such as \emph{"badass"} and \emph{"Suckit"}, it is not meant in a
hateful way. On the contrary, this tweet uses slang to express
admiration and support.

The 3rd example clearly is hateful, comparing the Democratic Party to
the Nazis and the position of Biden and Harris to Hindenburg and
Hitler. However, apparently the word Nazis is not sufficient to
communicate hate speech, while the other signals in this tweet are
presumably infrequent in the corpus as well.  These are interesting
examples which show that hate/offensive speech detection requires at
times world knowledge and common-sense reasoning (which BERT is
arguable only capable of to a very limited extent).

\subsection{Discussion}
The results in Table~\ref{stance_detection} show that the
disproportion among the classes \emph{Against, Favor, Neither, Mixed}
and \emph{Neutral mentions} seen in Tables~\ref{trump_hate}
and~\ref{biden_hate} are presumably influencing the performance. The
classes \emph{Mixed} and \emph{Neutral mentions} contain less tweets
than the other classes. Consequently, the model did not have the same
amount of training data for these two classes and tweets that should
be categorized as \emph{Neither} or \emph{Neutral} were
misclassified. In addition to \emph{Mixed} and \emph{Neutral
  mentions}, the class \emph{Against} of target Biden is also
outweighed by the dominant classes \emph{Favor} and \emph{Neither}
(see Table~\ref{biden_hate}).

When looking at the distribution of hateful and offensive and
non-hateful tweets, we see that our data set contains more non-hateful
tweets. As a result, the classification is biased. While
\citet{Davidson2017} created their data set with keywords from a hate
speech lexicon and \citet{Mandl2019} sampled their data with hashtags
and keywords for which hate speech can be expected, our data was
collected by using, but not limited to, offensive and hateful
mentions. Thus, our hate speech data is more imbalanced but provides
interesting insights into how people talk politics on Twitter. We
assume that our corpus exhibits a more realistic distribution of
hate/offensive speech for a particular topic than a subset of already
existing resources.

There may be some possible limitations in this study. Using Twitter as
data source provides challenges, because tweets contain noise,
spelling mistakes and incomplete sentences. Further, the specified
search criteria mentioned above might have had an effect on the
results. Next to the nicknames Trump uses for his opponents, most of
the keywords used to collect tweets refer to political
candidates. Mentions of the respective political parties such as
``Democrats'', ``Republicans'' etc. were not included in the
search. Yet, during the annotation we realized that it was not
possible to differentiate the candidates from their respective
parties. Hence, tweets were annotated for political parties and
candidates inferring from hashtags such as "\#VoteBlue2020" that the
tweeter argues in favor of Joe Biden.

\section{Conclusion and Future Work}
In this paper, we have investigated stance detection on political
tweets and whether or not supporters of Trump use more hate speech
than supporters of Biden (not significantly). We found that manual
annotation is possible with acceptable agreement scores, and that
automatic stance detection towards political candidates and parties is
possible with good performance.

The limitations of this study are twofold -- on the one side, future
work might want to consider to add the nicknames of all main
candidates and explicitly include social media posts about the party,
not only about the candidate, as we found a separation is often
difficult.  Further, we did not perform extensive hyperparameter
optimization in our neural approach.

We suggest that future work invests in developing computational models
that work across corpora and are able to adapt to domain and
time-specific as well as societal and situational expressions of hate
and offensive language. This is required, as our corpus shows that
some references to offensive content are realized by domain-specific
and societal expressions.

This might be realized by combining offensive language detection and
stance detection in a joint multi-task learning approach, potentially
including other aspects like personality traits or specific
emotions. We assume that such concepts can benefit from
representations in joint models.

\section*{Acknowledgements}
This project has been partially funded by Deutsche
Forschungsgemeinschaft (projects SEAT, KL 2869/1-1 and CEAT, KL
2869/1-2). We thank Anne Kreuter and Miquel Luj\'an for fruitful
discussions.\includegraphics{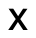}

\bibliographystyle{acl_natbib}
\bibliography{lit}

\begin{thebibliography}{28}
\expandafter\ifx\csname natexlab\endcsname\relax\def\natexlab#1{#1}\fi

\bibitem[{Aker et~al.(2017)Aker, Derczynski, and Bontcheva}]{Aker2017}
Ahmet Aker, Leon Derczynski, and Kalina Bontcheva. 2017.
\newblock \href {https://doi.org/10.26615/978-954-452-049-6_005} {Simple open
  stance classification for rumour analysis}.
\newblock In \emph{Proceedings of the International Conference Recent Advances
  in Natural Language Processing, {RANLP} 2017}, pages 31--39, Varna, Bulgaria.

\bibitem[{Basile et~al.(2019)Basile, Bosco, Fersini, Nozza, Patti,
  Rangel~Pardo, Rosso, and Sanguinetti}]{Basile2019}
Valerio Basile, Cristina Bosco, Elisabetta Fersini, Debora Nozza, Viviana
  Patti, Francisco~Manuel Rangel~Pardo, Paolo Rosso, and Manuela Sanguinetti.
  2019.
\newblock \href {https://doi.org/10.18653/v1/S19-2007} {{S}em{E}val-2019 task
  5: Multilingual detection of hate speech against immigrants and women in
  {T}witter}.
\newblock In \emph{Proceedings of the 13th International Workshop on Semantic
  Evaluation}, pages 54--63, Minneapolis, Minnesota, USA. Association for
  Computational Linguistics.

\bibitem[{Burnap and Williams(2014)}]{Burnap2014}
Peter Burnap and Matthew Williams. 2014.
\newblock Hate speech, machine classification and statistical modelling of
  information flows on twitter: interpretation and communication for policy
  decision making.
\newblock In \emph{Internet, Policy and Politics}, Oxford, United Kingdom.

\bibitem[{Clement(2019)}]{Clement2019}
Jessica Clement. 2019.
\newblock \href
  {https://www.statista.com/statistics/282087/number-of-monthly-active-twitter-users/}
  {Twitter: monthly active users worldwide}.
\newblock
  \url{https://www.statista.com/statistics/282087/number-of-monthly-active-twitter-users/}.

\bibitem[{Davidson et~al.(2017)Davidson, Warmsley, Macy, and
  Weber}]{Davidson2017}
Thomas Davidson, Dana Warmsley, Michael~W. Macy, and Ingmar Weber. 2017.
\newblock \href {https://aaai.org/ocs/index.php/ICWSM/ICWSM17/paper/view/15665}
  {Automated hate speech detection and the problem of offensive language}.
\newblock In \emph{Proceedings of the Eleventh International Conference on Web
  and Social Media, {ICWSM}}, pages 512--515, Montr{\'{e}}al, Qu{\'{e}}bec,
  Canada. {AAAI} Press.

\bibitem[{Devlin et~al.(2019)Devlin, Chang, Lee, and Toutanova}]{Devlin2019}
Jacob Devlin, Ming-Wei Chang, Kenton Lee, and Kristina Toutanova. 2019.
\newblock \href {https://doi.org/10.18653/v1/N19-1423} {{BERT}: Pre-training of
  deep bidirectional transformers for language understanding}.
\newblock In \emph{Proceedings of the 2019 Conference of the North {A}merican
  Chapter of the Association for Computational Linguistics: Human Language
  Technologies, Volume 1 (Long and Short Papers)}, pages 4171--4186,
  Minneapolis, Minnesota. Association for Computational Linguistics.

\bibitem[{Dinakar et~al.(2012)Dinakar, Jones, Havasi, Lieberman, and
  Picard}]{Dinakar2012}
Karthik Dinakar, Birago Jones, Catherine Havasi, Henry Lieberman, and Rosalind
  Picard. 2012.
\newblock \href {https://doi.org/10.1145/2362394.2362400} {Common sense
  reasoning for detection, prevention, and mitigation of cyberbullying}.
\newblock \emph{ACM Trans. Interact. Intell. Syst.}, 2(3).

\bibitem[{Fortuna and Nunes(2018)}]{Fortuna2018}
Paula Fortuna and S\'{e}rgio Nunes. 2018.
\newblock \href {https://doi.org/10.1145/3232676} {A survey on automatic
  detection of hate speech in text}.
\newblock \emph{ACM Comput. Surv.}, 51(4).

\bibitem[{Gao and Huang(2017)}]{GaoHuang2017}
Lei Gao and Ruihong Huang. 2017.
\newblock \href {https://doi.org/10.26615/978-954-452-049-6_036} {Detecting
  online hate speech using context aware models}.
\newblock In \emph{Proceedings of the International Conference Recent Advances
  in Natural Language Processing, {RANLP} 2017}, pages 260--266, Varna,
  Bulgaria. INCOMA Ltd.

\bibitem[{de~Gibert et~al.(2018)de~Gibert, Perez, Garc{\'\i}a-Pablos, and
  Cuadros}]{DeGibert2018}
Ona de~Gibert, Naiara Perez, Aitor Garc{\'\i}a-Pablos, and Montse Cuadros.
  2018.
\newblock \href {https://doi.org/10.18653/v1/W18-5102} {Hate speech dataset
  from a white supremacy forum}.
\newblock In \emph{Proceedings of the 2nd Workshop on Abusive Language Online
  ({ALW}2)}, pages 11--20, Brussels, Belgium. Association for Computational
  Linguistics.

\bibitem[{K{\"{u}}{\c{c}}{\"{u}}k and Can(2018)}]{Can2018}
Dilek K{\"{u}}{\c{c}}{\"{u}}k and Fazli Can. 2018.
\newblock \href {http://arxiv.org/abs/1803.08910} {Stance detection on tweets:
  An {SVM}-based approach}.
\newblock \emph{CoRR}, abs/1803.08910.

\bibitem[{Lozhnikov et~al.(2020)Lozhnikov, Derczynski, and
  Mazzara}]{Lozknikov2020}
Nikita Lozhnikov, Leon Derczynski, and Manuel Mazzara. 2020.
\newblock \href {https://doi.org/10.1007/978-3-030-14687-0_16} {Stance
  prediction for russian: Data and analysis}.
\newblock In \emph{Proceedings of 6th International Conference in Software
  Engineering for Defence Applications}, pages 176--186, Cham. Springer
  International Publishing.

\bibitem[{Mandl et~al.(2019)Mandl, Modha, Majumder, Patel, Dave, Mandlia, and
  Patel}]{Mandl2019}
Thomas Mandl, Sandip Modha, Prasenjit Majumder, Daksh Patel, Mohana Dave,
  Chintak Mandlia, and Aditya Patel. 2019.
\newblock \href {https://doi.org/10.1145/3368567.3368584} {Overview of the
  {HASOC} track at {FIRE} 2019: Hate speech and offensive content
  identification in indo-european languages}.
\newblock In \emph{Proceedings of the 11th Forum for Information Retrieval
  Evaluation}, FIRE '19, page 14–17, New York, NY, USA. Association for
  Computing Machinery.

\bibitem[{Mladenovi\'{c} et~al.(2021)Mladenovi\'{c}, O\v{s}mjanski, and
  Stankovi\'{c}}]{Mladenovic2021}
Miljana Mladenovi\'{c}, Vera O\v{s}mjanski, and Sta\v{s}a~Vuji\v{c}i\'{c}
  Stankovi\'{c}. 2021.
\newblock \href {https://doi.org/10.1145/3424246} {Cyber-aggression,
  cyberbullying, and cyber-grooming: A survey and research challenges}.
\newblock \emph{ACM Comput. Surv.}, 54(1).

\bibitem[{Mohammad et~al.(2016)Mohammad, Kiritchenko, Sobhani, Zhu, and
  Cherry}]{Mohammad2016}
Saif Mohammad, Svetlana Kiritchenko, Parinaz Sobhani, Xiaodan Zhu, and Colin
  Cherry. 2016.
\newblock \href {https://doi.org/10.18653/v1/S16-1003} {{S}em{E}val-2016 task
  6: Detecting stance in tweets}.
\newblock In \emph{Proceedings of the 10th International Workshop on Semantic
  Evaluation ({S}em{E}val-2016)}, pages 31--41, San Diego, California.
  Association for Computational Linguistics.

\bibitem[{Mohammad et~al.(2017)Mohammad, Sobhani, and
  Kiritchenko}]{Mohammad2017}
Saif~M. Mohammad, Parinaz Sobhani, and Svetlana Kiritchenko. 2017.
\newblock \href {https://doi.org/10.1145/3003433} {Stance and sentiment in
  tweets}.
\newblock \emph{ACM Trans. Internet Technol.}, 17(3).

\bibitem[{Nockelby(2000)}]{Nockelby2000}
John~T. Nockelby. 2000.
\newblock Hate speech.
\newblock \emph{Encyclopedia of the American Constitution}, 3:1277–1279.

\bibitem[{Roß et~al.(2016)Roß, Rist, Carbonell, Cabrera, Kurowsky, and
  Wojatzki}]{Ross2016}
Björn Roß, Michael Rist, Guillermo Carbonell, Benjamin Cabrera, Nils
  Kurowsky, and Michael~Maximilian Wojatzki. 2016.
\newblock \href {https://doi.org/10.17185/duepublico/42132} {Measuring the
  reliability of hate speech annotations: The case of the european refugee
  crisis}.
\newblock
  \url{https://duepublico2.uni-due.de/receive/duepublico_mods_00042132}.

\bibitem[{Schmidt and Wiegand(2017)}]{Schmidt2017}
Anna Schmidt and Michael Wiegand. 2017.
\newblock \href {https://doi.org/10.18653/v1/W17-1101} {A survey on hate speech
  detection using natural language processing}.
\newblock In \emph{Proceedings of the Fifth International Workshop on Natural
  Language Processing for Social Media}, pages 1--10, Valencia, Spain.
  Association for Computational Linguistics.

\bibitem[{Somasundaran and Wiebe(2010)}]{Somasundaran2010}
Swapna Somasundaran and Janyce Wiebe. 2010.
\newblock \href {https://www.aclweb.org/anthology/W10-0214} {Recognizing
  stances in ideological on-line debates}.
\newblock In \emph{Proceedings of the {NAACL} {HLT} 2010 Workshop on
  Computational Approaches to Analysis and Generation of Emotion in Text},
  pages 116--124, Los Angeles, CA. Association for Computational Linguistics.

\bibitem[{Spertus(1997)}]{Spertus1997}
Ellen Spertus. 1997.
\newblock Smokey: Automatic recognition of hostile messages.
\newblock In \emph{Proceedings of the Fourteenth National Conference on
  Artificial Intelligence and Ninth Conference on Innovative Applications of
  Artificial Intelligence}, AAAI'97/IAAI'97, page 1058–1065. AAAI Press.

\bibitem[{Stolee and Caton(2018)}]{Stolee2018}
Galen Stolee and Steve Caton. 2018.
\newblock \href {https://doi.org/10.1086/694755} {Twitter, trump, and the base:
  A shift to a new form of presidential talk?}
\newblock \emph{Signs and Society}, 6:147--165.

\bibitem[{Thomas et~al.(2006)Thomas, Pang, and Lee}]{Thomas2006}
Matt Thomas, Bo~Pang, and Lillian Lee. 2006.
\newblock \href {https://www.aclweb.org/anthology/W06-1639} {Get out the vote:
  Determining support or opposition from congressional floor-debate
  transcripts}.
\newblock In \emph{Proceedings of the 2006 Conference on Empirical Methods in
  Natural Language Processing}, pages 327--335, Sydney, Australia. Association
  for Computational Linguistics.

\bibitem[{Tumasjan et~al.(2010)Tumasjan, Sprenger, Sandner, and
  Welpe}]{Tumasjan2010}
Andranik Tumasjan, Timm Sprenger, Philipp Sandner, and Isabell Welpe. 2010.
\newblock Predicting elections with twitter: What 140 characters reveal about
  political sentiment.
\newblock In \emph{Fourth International AAAI Conference on Weblogs and Social
  Media}, pages 178--185.

\bibitem[{Warner and Hirschberg(2012)}]{Warner2012}
William Warner and Julia Hirschberg. 2012.
\newblock \href {https://www.aclweb.org/anthology/W12-2103} {Detecting hate
  speech on the world wide web}.
\newblock In \emph{Proceedings of the Second Workshop on Language in Social
  Media}, pages 19--26, Montr{\'e}al, Canada. Association for Computational
  Linguistics.

\bibitem[{Waseem and Hovy(2016)}]{Waseem2016}
Zeerak Waseem and Dirk Hovy. 2016.
\newblock \href {https://doi.org/10.18653/v1/N16-2013} {Hateful symbols or
  hateful people? predictive features for hate speech detection on {T}witter}.
\newblock In \emph{Proceedings of the {NAACL} Student Research Workshop}, pages
  88--93, San Diego, California. Association for Computational Linguistics.

\bibitem[{Wolf et~al.(2020)Wolf, Debut, Sanh, Chaumond, Delangue, Moi, Cistac,
  Rault, Louf, Funtowicz, Davison, Shleifer, von Platen, Ma, Jernite, Plu, Xu,
  Scao, Gugger, Drame, Lhoest, and Rush}]{Wolf2020}
Thomas Wolf, Lysandre Debut, Victor Sanh, Julien Chaumond, Clement Delangue,
  Anthony Moi, Pierric Cistac, Tim Rault, Rémi Louf, Morgan Funtowicz, Joe
  Davison, Sam Shleifer, Patrick von Platen, Clara Ma, Yacine Jernite, Julien
  Plu, Canwen Xu, Teven~Le Scao, Sylvain Gugger, Mariama Drame, Quentin Lhoest,
  and Alexander~M. Rush. 2020.
\newblock \href {https://www.aclweb.org/anthology/2020.emnlp-demos.6}
  {Transformers: State-of-the-art natural language processing}.
\newblock In \emph{Proceedings of the 2020 Conference on Empirical Methods in
  Natural Language Processing: System Demonstrations}, pages 38--45, Online.
  Association for Computational Linguistics.

\bibitem[{Zampieri et~al.(2020)Zampieri, Nakov, Rosenthal, Atanasova,
  Karadzhov, Mubarak, Derczynski, Pitenis, and
  {\c{C}}{\"o}ltekin}]{Zampieri2020}
Marcos Zampieri, Preslav Nakov, Sara Rosenthal, Pepa Atanasova, Georgi
  Karadzhov, Hamdy Mubarak, Leon Derczynski, Zeses Pitenis, and
  {\c{C}}a{\u{g}}r{\i} {\c{C}}{\"o}ltekin. 2020.
\newblock \href {https://www.aclweb.org/anthology/2020.semeval-1.188}
  {{S}em{E}val-2020 task 12: Multilingual offensive language identification in
  social media ({O}ffens{E}val 2020)}.
\newblock In \emph{Proceedings of the Fourteenth Workshop on Semantic
  Evaluation}, pages 1425--1447, Barcelona (online). International Committee
  for Computational Linguistics.

\end{thebibliography}

\end{document}